\pdfoutput=1

\documentclass[11pt]{article}

\usepackage{ACL2023}

\usepackage{times}
\usepackage{latexsym}

\usepackage[T1]{fontenc}

\usepackage[utf8]{inputenc}

\usepackage{microtype}

\usepackage{inconsolata}

\usepackage{hyperref}
\usepackage{url}

\usepackage{algorithm}
\usepackage[algo2e, linesnumbered]{algorithm2e}

\usepackage{listings}
\usepackage{color}
\usepackage{amsmath}
\usepackage{amssymb}
\usepackage{mathtools}
\usepackage{amsthm}
\usepackage{bm}
\usepackage{multirow}
\usepackage{subfigure}
\usepackage{booktabs}
\usepackage{wrapfig}
\usepackage{adjustbox}
\usepackage{threeparttable}
\usepackage{graphicx}
\usepackage{subcaption}
\theoremstyle{plain}

\theoremstyle{definition}

\theoremstyle{remark}

\newcommand{\bs}[1]{\boldsymbol{#1}}

\title{On Diversified Preferences of Large Language Model Alignment}

\author{Dun Zeng$^{1,}$\footnotemark[1]\footnotemark[3]$^{\;\,}$ \quad YongDai$^{2,}$\footnotemark[1]\footnotemark[3] \quad Pengyu Cheng$^{1,}$\footnotemark[1]$^{\;\,}$ \quad \textbf{Longyue Wang}$^{3,}$$\footnotemark[3]$ \\ \textbf{Tianhao Hu}$^{1}$ \quad \textbf{Wanshun Chen}$^{1}$ \quad \textbf{Nan Du}$^{1}$ \quad \textbf{Zenglin Xu}$^{4,}$\footnotemark[2] \\
$^{1}$Tencent AI Lab, $^{2}$HiThink Research, $^{3}$Alibaba Group, $^{4}$Peng Cheng Lab  \\
}

\begin{document}
\maketitle

\renewcommand*{\thefootnote}{\fnsymbol{footnote}}
\footnotetext[1]{Equal contribution.}
\footnotetext[2]{Corresponding author: zenglin@gmail.com}
\footnotetext[3]{Part of work was done at Tencent AI Lab}

\renewcommand*{\thefootnote}{\arabic{footnote}}

\begin{abstract}
Aligning large language models (LLMs) with human preferences has been recognized as the key to improving LLMs' interaction quality. However, in this pluralistic world, human preferences can be diversified due to annotators' different tastes, which hinders the effectiveness of LLM alignment methods. 
This paper presents the first quantitative analysis of the experimental scaling law for reward models with varying sizes, from 1.3 billion to 7 billion parameters, trained with human feedback exhibiting diverse preferences.
Our analysis reveals that the impact of diversified human preferences depends on both model size and data size. Larger models with sufficient capacity mitigate the negative effects of diverse preferences, while smaller models struggle to accommodate them. To mitigate the impact of diverse preferences, we introduce a new metric, Expected Calibration Error (ECE), to evaluate RMs and show their obvious positive correlation with the alignment performance of LLMs.
Furthermore, we propose a Multi-Objective Reward learning method (MORE) to enhance the calibration performance of RMs on shared preferences. 
Through experiments on four models and five human preference datasets, we find the calibration error can be adopted as a key metric for evaluating RMs and MORE can obtain superior alignment performance.
\end{abstract}

\section{Introduction}\label{sec:introduction}
 Large language models (LLMs), such as ChatGPT~\cite{chatgpt} and LLaMa~\citep{touvron2023LLaMa,touvron2023LLaMa2}, have significantly accelerated the development process toward artificial general intelligence (AGI). 
 Among the key factors for such great achievement, the \textit{alignment} technique,  which finetunes LLMs with human feedback~\citep{christiano2017deep}, has played an essential role in training LLMs' responses to follow human values (\textit{e.g.}, helpfulness and harmlessness)~\citep{askell2021general}.
 Among the LLM alignment algorithms, reinforcement learning from human feedback (RLHF)~\citep{ouyang2022training} has become the mainstream solution, which first learns a reward model (RM) representing human preferences and then updates LLMs via the proximal policy optimization (PPO)~\citep{schulman2017proximal} toward generating responses with higher RM scores. 
Alternative alignment methods also have been sequentially proposed for better computational complexity and training instability, such as 
RAFT~\cite{dong2023raft},
DPO~\cite{rafailov2023direct}, RRHF~\cite{yuan2023rrhf}, and APO~\cite{cheng2023adversarial}. 
%

%

The performance of these alignment methods highly depends on the quality of human preference data $(\bs{x}, \bs{y}_w, \bs{y}_l)$, where $\bs{x}$ is the input query to the LLM, and response $\bs{y}_w$ is preferred to response $\bs{y}_l$ under the human annotation~\cite{ouyang2022training}. Ideally, the preference datasets should \textit{uniformly} be helpful, harmless, benevolent, and unbiased to guide the LLM alignment. However, in real-world scenarios, individuals can have \textit{diversified} preferences on the same topic based on their different experiences, educational backgrounds, religions, and cultures~\citep{DBLP:conf/emnlp/LeonardelliMAGT21}.
Even for the same person, his or her expected model answer to a particular question can vary depending on different  scenarios~\citep{cheng2023everyone}. 
The annotation disagreement, which is caused by different annotators or the same annotator in different scenarios~\cite{bai2022training}, will significantly hinder the effectiveness of alignment methods~\cite{davani2022dealing, wan2023everyone, he2024shed}. 


  
%

\begin{figure*}[t]
\vspace{-6mm}
\centering
\includegraphics[width=0.9\linewidth]{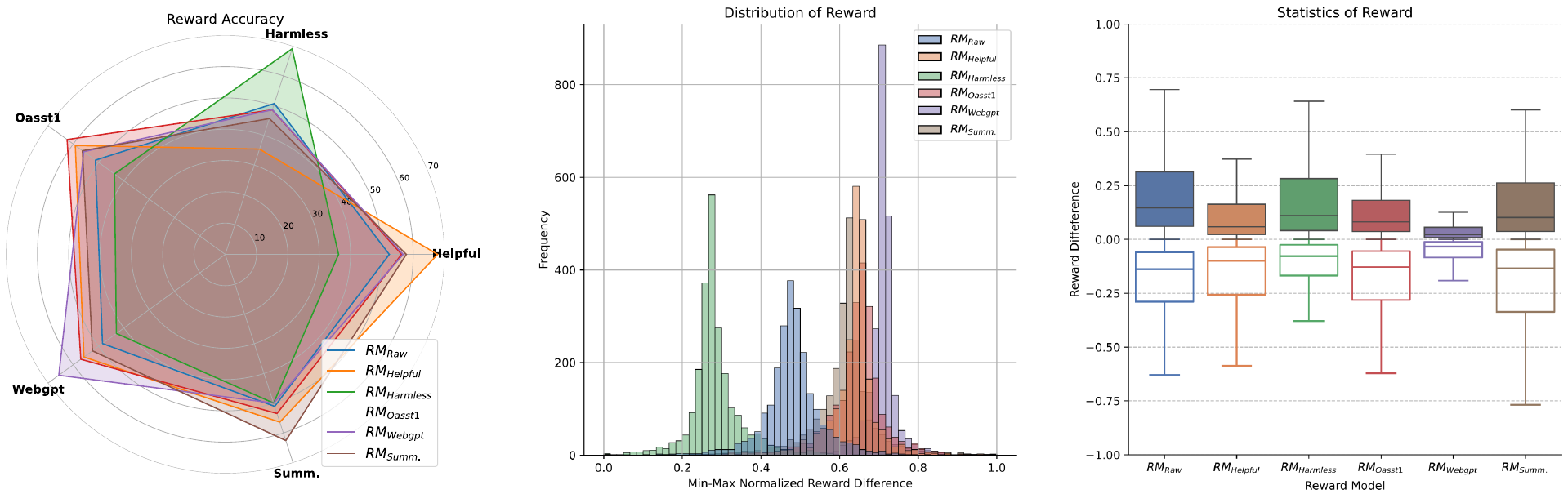}
\vspace{-2mm}
\caption{Illustration of \textit{Diversified Preferences}. \textbf{Left:} reward accuracy on each preference. \textbf{Middle:} the reward distribution of each RM on harmless preference. \textbf{Right:} the reward statistics of each RM on harmless preference. The solid box indicates the reward statistics on correct rewarded samples, and the hollow box indicates the wrong rewarded samples. }\label{fig:observation}
\vspace{-5mm}
\end{figure*}

To identify the diversified preferences quantitatively, we select five commonly used human feedback datasets, train an RM on each, and then test the performance on the other sets (details in Section~\ref{sec:preniminary}). We plot the observation results in Figure~\ref{fig:observation}. We observe that training RM on a single preference data source may cause inconsistent reward distribution shifts (middle plot), result in diverse reward values (right plot), and compromise the performance of other sets (left plot). The result indicates that different human preference datasets have different preference distributions~\citep{cheng2023everyone}. 
Hence, a more comprehensive understanding of the impact of diversified human preference datasets on the reward model becomes crucial, yet it has not received adequate attention and remained unexplored in the LLM alignment domain.

In our exploration, we found the \textit{over-rewarding} phenomenon, that is, the vanilla RMs tend to output extreme rewards on samples, which damages the RMs and LLM alignment. To enhance the efficiency of leveraging the diversified preference datasets, inspired by multi-objective optimization methods~\cite{sener2018multi, zeng2023tackling}, we regard RMs as a \textit{shared reward} additionally with a customized \textit{reward drift}. The shared reward represents the shared preferences across datasets (or general human preferences) and the reward drift contains individual or domain-specific preference information~\citep{cheng2023everyone}. 
Then, we introduce a \textbf{M}ulti-\textbf{O}bjective \textbf{Re}ward training scheme (MORE) to capture the shared (general) preference information, which adopts a novel reweight techniques to minimize the mean gradient of enlarging reward drifts. With MORE, RMs can capture a broader range of preferences and mitigate the impact of reward drifts. The main contributions of this paper are: 
\begin{itemize}
\item  This is the first work to demonstrate the positive correlation between the calibration performance of RMs and the alignment performance of LLMs.
Moreover, RM learning on diversified preferences typically induces high calibration errors, indicating unreliable rewards. The unreliable rewards come from a \textit{over-rewarding} phenomenon, denoting vanilla RMs output extreme rewards inducing harmful \textit{reward drifts}. Hence, it negatively impacts the performance of LLM alignment.

\item  We induce a simple and effective Multi-Objective Reward (MORE) training scheme to alleviate the over-rewarding phenomenon. MORE makes self-adaption to the RM learning gradient to mitigate the \textit{reward drifts}. MORE effectively enhances the calibration performance of RMs, especially on shared preferences across diversified preference datasets.

\item  We verified our findings with Pythia-1.4B, Pythia-2.8B~\citep{biderman2023pythia} and LLaMa2-7B~\citep{touvron2023LLaMa2} on \textbf{five} widely recognized and diverse preference datasets. 
Through empirical analysis, we established that MORE significantly minimizes reward drift and achieves low \textit{Expected Calibration Error} (ECE) values. Additionally, by applying reject sampling to Alpaca-7B~\citep{alpaca} with the RMs generated, we aligned the models with \textit{Helpful\&Harmless} preferences, thereby affirming the critical role of ECE in the evaluation of Reward Models.
\end{itemize}

\section{Background}
\paragraph{Large language Model Alignment}
Parameterized by $\bs{\theta}$, a reward model (RM) is a mapping $r_{\bs{\theta}}: \mathcal{X} \times \mathcal{Y} \rightarrow \mathbb{R}$, which provides a real-valued reward score $r_{\bs{\theta}}(\bs{x},\bs{y})$ evaluating a textual response $\bs{y} = (\bs{y}_1, \bs{y}_2, \dots, \bs{y}_M) \in \mathcal{Y}$ corresponding to an input prompt $\bs{x}=(\bs{x}_1, \bs{x}_2, \dots, \bs{x}_N) \in \mathcal{X}$. 
Given a sample $(\bs{x}, \bs{y}_w, \bs{y}_l) \sim \mathcal{D}$ from  a preference dataset $\mathcal{D}$, $r_{\bs{\theta}}$ is expected to provide a preference score with $r_{\bs{\theta}}(\bs{x}, \bs{y}_w) > r_{\bs{\theta}}(\bs{x}, \bs{y}_l)$, representing the response $\bs{y}_w$ is preferred. Following the Bradley-Terry model~\citep{david1963method}, the RM learning objective on the preference dataset $(\bs{x}, \bs{y}_{w}, \bs{y}_{l}) \sim \mathcal{D}$ is defined as:
\begin{equation}\label{eq:vanilla_rankloss} \textstyle
\mathcal{L}_{\text{rank}}(\bs{\theta}; \mathcal{D}) = - \mathbb{E}_{\mathcal{D}} \left[\log(\sigma\left(\Delta r_{\bs{\theta}}(\bs{y}_w, \bs{y}_l)\right))\right]
\end{equation}
where we use $\Delta r_{\bs{\theta}}(\bs{y}_w, \bs{y}_l)$ to denote reward difference $r_{\bs{\theta}}(\bs{x},\bs{y}_w)-r_{\bs{\theta}}(\bs{x},\bs{y}_l)$ for simplifying notation in this paper and $\sigma(\cdot)$ is the Sigmoid function. With a well-learned reward $r_{\bs{\theta}}(\bs{x},\bs{y})$, LLM alignment optimizes the generation policy $\pi(\bs{y}|\bs{x})$ by maximizing the expected reward value:
%
\begin{equation}\label{eq:rlhf}
\begin{aligned}
&\mathbb{E}_{\bs{x}\sim \mathcal{D}, \bs{y}\sim \pi(\bs{y}|\bs{x})} [r_{\bs{\theta}}(\bs{x},\bs{y})]  \\
&\quad - \beta \mathbb{D}_{\text{KL}}[\pi(\bs{y}|\bs{x})\| \pi_{\text{ref}}(\bs{y}|\bs{x})],
\end{aligned}
\end{equation}
where $\mathbb{D}_{\text{KL}}[\pi(\bs{y}|\bs{x})\| \pi_{\text{ref}}(\bs{y}|\bs{x})]$ is the KL divergence regularizer between current policy $\pi$ and a reference $\pi_\text{ref}$, preventing the optimization from instability and degeneration. The typical solution to the preference optimization in equation 3 is reinforcement learning (RLHF)~\citep{ouyang2022training}, especially with the proximal policy optimization (PPO) algorithms~\citep{schulman2017proximal}. However, RLHF has been recognized as practically suffering from implementation complexity and training instability. To avoid the RL schedule during alignment, reject sampling methods~\citep{liu2023statistical} directly conduct supervised fine-tuning on $\bs{y}^{\text{best}}$ to further simplify the human preference alignment process. The rejection sampling optimization (RJS) loss can be written as
\begin{equation}\label{eq:rjs}
    \mathcal{L}_{\text{RJS}}(\pi) = - \mathbb{E}_{\bs{x}\sim \mathcal{D}, \bs{y}\sim \pi(\bs{y}|\bs{x})} [\log \pi(\bs{y}^{\text{best}} | \bs{x})],
\end{equation}
where $\bs{y}^{\text{best}} = \arg \max_{1\leq s \leq S\{r(\bs{x},\bs{y}^s)\}}$ is the sampled response with the highest reward score.

\paragraph{Calibration Error} 
Calibration error is an effective method to estimate the confidence of a model's outputs~\citep{guo2017calibration}. We divide the confidence interval $[0,1]$ with finite samples into $M$ bins with equal length ($1/M$). Then, we place model predictions into these bins according to their prediction confidence. Let $B_m$ be the set of indices of samples that fall into the internal $(\frac{m-1}{M}, \frac{m}{M}]$. We calculate the corresponding accuracy and average confidence of each bin as follows:
$$
\begin{aligned}
& \operatorname{Acc}\left(B_m\right)  =\frac{1}{\left|B_m\right|} \sum_{i \in B_m} \mathbb{I}\left(\hat{y}_i=y_i\right),  \\
& \operatorname{Conf}\left(B_m\right)  =\frac{1}{\left|B_m\right|} \sum_{i \in B_m} \hat{p}_i,
\end{aligned}
$$
where $\hat{y}_i$ are the prediction results, and $y_i$ is the ground-truth of the $i$-th sample. $\mathbb{I}$ is the indicator function which produces 1 if $\hat{y}_i = y_i$ otherwise 0. $\hat{p}_i$ is the prediction confidence of the $i$-th sample. In the context of reward modeling, the prediction confidence $\hat{p}_i = \sigma(\cdot)$ in \eqref{eq:vanilla_rankloss}. For a set of $N$ samples, we can compute the \textit{Expected Calibration Error} as follows:
$$
ECE = \sum_{m=1}^M \frac{\left|B_m\right|}{N}\left|\operatorname{Acc}\left(B_m\right)-\operatorname{Conf}\left(B_m\right)\right|.
$$
We set $M=10$ for measuring calibration performance in this paper. 

Numerous studies have focused on improving the calibration performance of statistical machine-learning systems~\citep{degroot1983comparison, palmer2008toward, yang2010nurses}. Furthermore, the calibration error of neural networks provides additional information for users to determine whether to trust the model's predictions, especially for modern neural networks that are more challenging to interpret~\citep{guo2017calibration, zhu2023calibration}. In the field of natural language processing, studies have revealed a positive relationship between calibration performance and the reduction of hallucination~\citep{xiao2021hallucination, tian2019sticking}, and the evaluation of pre-trained language models~\citep{kadavath2022language, tian2023just}. The calibration error has demonstrated its ability to evaluate the performance of language models. In this paper, we first employ the calibration error to evaluate the RMs. Subsequently, we investigate the implicit connection between RMs and LLM alignment under diversified preferences.

\section{Empirical Study of Diversified Preferences}\label{sec:preniminary}

We start with an empirical analysis of diversified preferences in reward modeling on multiple sources $\mathcal{D}=\{\mathcal{D}_1, \dots, \mathcal{D}_K\}$, where each data source $D_k$ contains the preference comparison pairs from different tasks~\citep{dong2023abilities}, domains~\citep{cheng2023everyone}, or individuals~\cite{bai2022training}.  
In this paper, we selected \textit{Summarize}~\citep{stienon2020learning},
\textit{Webgpt}~\citep{DBLP:journals/corr/abs-2112-09332},
\textit{Helpful\&Harmless}~\citep{bai2022training},  and \textit{OASST1}~\citep{DBLP:journals/corr/abs-2304-07327} as the different preference sources to empirical analysis the phenomena of diversified preferences. We use Pythia-1.4B~\citep{biderman2023pythia} as the RM base, and finetuned RMs with comparisons from each source. The experiment setup aligns with Section~\ref{sec:rm_exp}.

The reward distributions across various RMs exhibit diversity when applied to the same dataset. We analyze and present the variation in rewards (defined as the difference in reward values assigned by an RM to the winning and losing samples) offered by these RMs, as illustrated in Figure~\ref{fig:observation} (additional results in Figure~\ref{fig:reward_dist} and~\ref{fig:additional-results} in Appendix). Compared with the results of raw model $\text{RM}_{\text{Raw}}$, we observe that training on different datasets results in diverse reward values (right plot) and distribution shift (middle plot). Specifically, the reward value distribution of $\text{RM}_{\text{Harmless}}$ shifts from the $\text{RM}_{\text{Raw}}$ in a certain degree. While the reward value distributions of $\text{RM}_{\text{Helpful}}$, $\text{RM}_{\text{Webgpt}}$, $\text{RM}_{\text{Oasst1}}$ and $\text{RM}_{\text{Summ.}}$ shifts to the a different direction. Moreover, despite the distribution of $\text{RM}_{\text{Helpful}}$, $\text{RM}_{\text{Webgpt}}$, $\text{RM}_{\text{Oasst1}}$ and $\text{RM}_{\text{Summ.}}$ are similar, the mean-variance of their reward values are quite different.

Furthermore, when considering the accuracy gains illustrated in Figure~\ref{fig:observation} (left plot), the observed shift in reward distribution indicates that the learned reward values from preference datasets are diversified. To effectively capture the shared reward values across these diversified preferences, it becomes necessary to formulate a new problem approach for reward modeling on diverse preference datasets.

\vspace{-2mm}
\section{Multi-Objective Reward Learning}

In this section, we propose our reward modeling on diversified preference datasets, highlighting the implicit reward drift during the reward learning process and its negative impacts. Then, we present the MORE training schemes to mitigate the reward drifts as a feasible solution. To maintain the integrity of our paper, we leave our quantitative analyses of reward modeling on diversified preferences in the next section.

\subsection{Preference Diversity as Reward Drift} 
We denote $r^*(\cdot,\cdot)$ as the shared reward function, which (ideally) provides reward values reflecting the general values among people (or shared preference information across datasets in practice). As the collected human-feedback datasets are limited and implicitly biased, training an RM $r_{\bs{\theta}}$ on a limited preference dataset can be viewed as drifting from an optimal reward. We can form a reward model $r_{\bs{\theta}}(\cdot, \cdot)$ with reward drift in a data level:
\begin{equation}\label{eq:reward_drift}
r_{\bs{\theta}}(\bs{x},\bs{y}) = r_{\bs{\theta}}^*(\bs{x},\bs{y}) + \tilde{r}_{\bs{\theta}}(\bs{x},\bs{y}),
\end{equation}
where $\bs{x},\bs{y} \in \mathcal{X} \times \mathcal{Y}$, and $\tilde{r}_{\bs{\theta}}(\bs{x},\bs{y})$ is the reward drift learned by RM $r_{\bs{\theta}}(\cdot, \cdot)$. 
Then, we investigate the vanilla ranking loss for reward modeling. Substituting reward function in \eqref{eq:vanilla_rankloss} with the drifted form \eqref{eq:reward_drift}, we have $\mathcal{L}_{\text{rank}}(\bs{\theta}; \mathcal{D})=$
\begin{align}\label{eq:drift_loss}
- \mathbb{E}_{\mathcal{D}} [ \log(\sigma(\Delta r_{\bs{\theta}}^*(\bs{y}_w, \bs{y}_l) + \Delta \tilde{r}_{\bs{\theta}}(\bs{y}_w, \bs{y}_l) ))]. 
\end{align}
Hence, updating the RM to minimize the rank loss will enlarge the reward differences (input of the Sigmoid function). Simultaneously, the reward drift is also enlarged, causing over-rewarding.


\begin{figure*}[t]
\centering
\vspace{-6mm}
\includegraphics[width=0.95\linewidth]{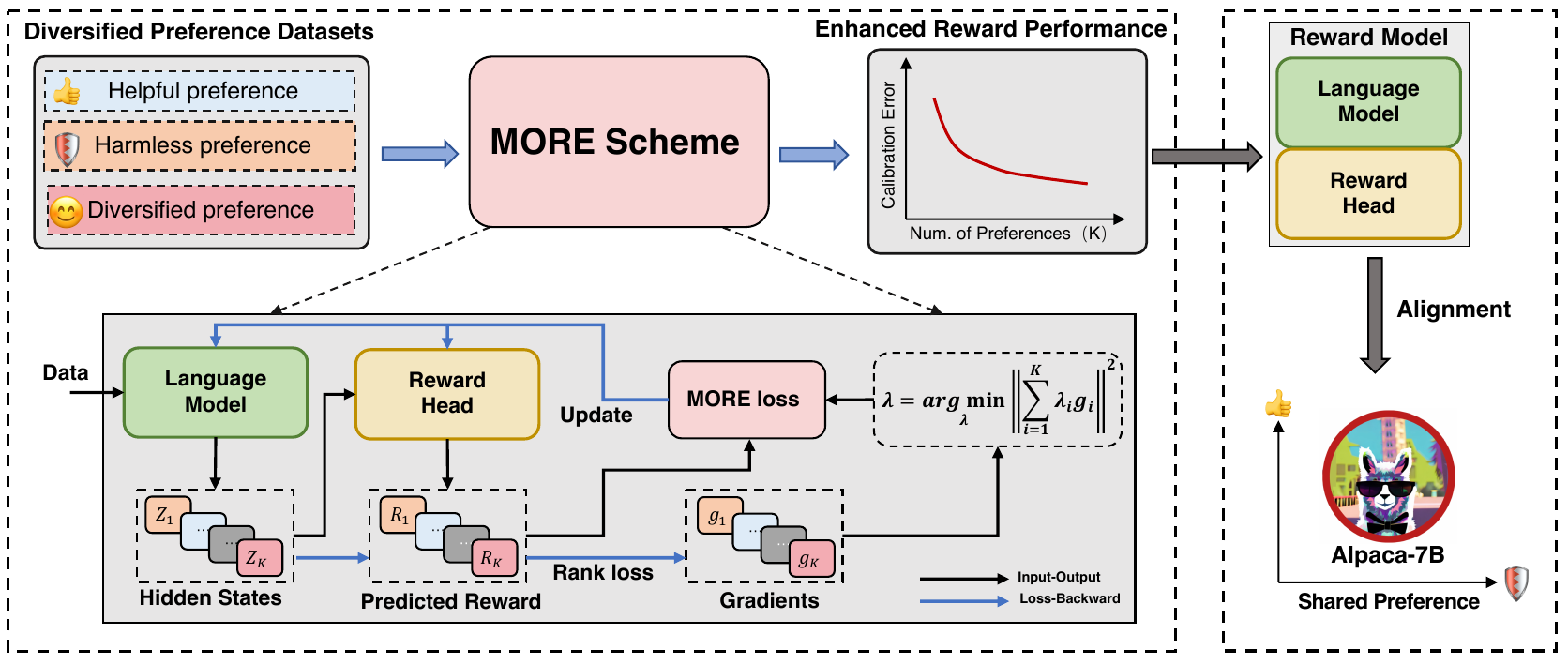}
 \vspace{-2mm}
\caption{Multi-objective reward model training scheme (MORE), which 
consists of four steps: (1) collect a diversified batch of data from the mixed dataset; (2) calculate the RM gradient for each preference source; (3) minimize the reward drift to determine the scalar $(\lambda_1, \lambda_2, \dots, \lambda_K)$ for MORE loss; 
 (4) update the RM with the re-weighted RM loss. Lower calibration error indicates the RM provides an accurate reward.}\label{fig:overview}
 \vspace{-4mm}
\end{figure*}

\subsection{Reward Modeling on Diversified Data} 



Letting $\bs{\theta}$ be the RM trained on mixed diverse datasets $\mathcal{D} = \{\mathcal{D}_1, \dots, \mathcal{D}_K\}$, the $r_{\bs{\theta}}(\bs{x},\bs{y})$ can be viewed as a multi-task learner with shared parameters~\citep{sener2018multi}. Then, the reward value provided by $r_{\bs{\theta}}(\bs{x},\bs{y})$ can be decomposed into voting format weighted by an implicit $\lambda$:
\begin{equation}\label{eq:biased_reward}
\begin{aligned}
\textstyle
r_{\bs{\theta}}(\bs{x},\bs{y}) =  r_{\bs{\theta}}^*(\bs{x},\bs{y}) + \sum_{i=1}^K \lambda_i \tilde{r}_{\bs{\theta}_i}(\bs{x},\bs{y}),
\end{aligned}
\end{equation}
where the shared reward $r^*_{\bs{\theta}} (\cdot,\cdot)$ is the same with arbitrary $\lambda$, and $\tilde{r}_{\bs{\theta}_i}(\cdot,\cdot)$ is the reward drift. We interpret that the $\tilde{r}_{\bs{\theta}_i}(\cdot,\cdot)$ is provided by subset of parameters $\bs{\theta}_i$, representing the preferences from the $i$-th dataset $\mathcal{D}_i$. This reward value decomposition naturally holds in the model output level, despite the non-linear nature of neural networks. 

Moreover, our formulation aligns with multi-task learning~\citep{crawshaw2020multi} and multi-objective learning~\citep{guardieiro2023multi} problems. For example, the $\bs{\theta}$ can be implemented as an ensemble model, where $\{{\bs{\theta}_i}\}, i\in [N]$ is the base models. Therefore, it is natural to adjust the weight $\lambda$ in an ensemble manner~\citep{coste2023reward, jang2023personalized, touvron2023LLaMa, eisenstein2023helping} to mitigate the reward drift such that $\min \sum_{i=1}^K \lambda_i \tilde{r}_{\bs{\theta}_i}(\bs{x},\bs{y})$. Compared with average rewards from multiple RMs~\cite{jang2023personalized, eisenstein2023helping}, we focus on training a single RM that learns the shared preference. We propose to reduce the model update on reward drift during RM training via linear scalarization~\cite{barrett2008learning}. Moreover, we provide further discussion on related manners in Section~\ref{sec:discussion}.



\subsection{Training Scheme: MORE} 

\paragraph{MORE loss function} 
Our analyses suggest finding proper weights $\lambda $ for mitigating reward drifts. Then, we propose training RMs to capture the shared preference across multiple datasets with the following objective: 
\begin{equation}\label{eq:r_loss}
\begin{aligned}\textstyle
\mathcal{L}_{\text{MORE}}(\bs{\theta}; \mathcal{D}) = \sum_{i=1}^K \lambda_i \mathcal{L}_{\text{rank}}(\bs{\theta}, \mathcal{D}_i),
\end{aligned}
\end{equation}
where $\sum_i^K \lambda_i = 1, \lambda_i \geq 0$. Compared with vanilla ranking loss in \eqref{eq:vanilla_rankloss}, the above loss additionally focuses on the combination relation across preferences. The linear combination of loss functions is commonly adopted in deep learning methods to balance the interaction of different modules~\citep{zhang2023blind, kurin2022defense}. Analogously, we treat each preference as an individual module and balance them wisely. Moreover, this formulation also covers several typical training cases. For example, directly mixing diverse preference datasets $\mathcal{D} = \{\mathcal{D}_1, \dots, \mathcal{D}_K\}$ and training a RM implicitly induces $\lambda_i = |\mathcal{D}_i|/|\mathcal{D}|$~\citep{mcmahan2017communication, rame2024warm}. Therefore, if the number of data samples from a single preference is greatly larger than other preferences, the RM is likely to drift to the preference with more samples. Excluding data quantity, the weight is also decided by the quality of data samples in the training process~\citep{katharopoulos2018not, zhou2023samples}. Neural network training typically provides a larger gradient for harder samples~\citep{katharopoulos2018not}, therefore, leaning the RMs preferences drift to these hard samples. In practice, the quantity and quality variance in diversified datasets may require more hyper-parameter searching~\citep{guo2024controllable} or data composition efforts~\citep{dong2023abilities} in the vanilla finetuning process. 

\paragraph{What is MORE doing?} 
We suggest training a better RM via self-adaption training weights $\lambda$ for better data efficiency. The MORE loss minimizes the ranking loss by solving a reward drift mitigation task, applying a \textit{batch-wise reweighting} method. Let batch data $\mathcal{B} = \{\bs{x}^{(b)}, \bs{y}_w^{(b)}, \bs{y}_l^{(b)}\}_{b=1}^B \sim \mathcal{D}$ be the sampled batch data from diverse datasets. Furthermore, $\mathcal{B}_i \sim \mathcal{D}_i \subset \mathcal{B}, \forall i\in[K]$ is the subset of batch data from the $i$-th preference dataset. 
We have the gradient $\nabla_{\bs{\theta}} \mathcal{L}_{\text{MORE}}(\bs{\theta}; \mathcal{B})$
\begin{equation}\label{eq:expand_more}
\begin{aligned}
& = \sum_{b=1}^B \left[-\nabla_{\bs{\theta}} \log(\sigma(\Delta r_{\bs{\theta}}^*(\bs{y}_w^{(b)},\bs{y}_l^{(b)}))) \right] + K \cdot \\
&\underbrace{\min \sum_{i=1}^K \lambda_i \sum_{j=1}^{|\mathcal{B}_i|} \left[-\nabla_{\bs{\theta}} 
 \log(\sigma(\Delta \tilde{r}_{\bs{\theta}}(\bs{y}_w^{(j)},\bs{y}_l^{(j)}))\right]}_{\text{Reward Drift Mitigation}},
\end{aligned}
\end{equation}
where we adjust $\lambda$ to minimize the partial gradient of enlarging reward drifts. The mitigation task in \eqref{eq:expand_more} can be efficiently solved by the Frank-Wolfe solver~\citep{DBLP:conf/icml/Jaggi13, sener2018multi, zhou2022convergence, zeng2023tackling}. We provide the details of our efficient implementation in the Appendix~\ref{sec:appendix}. Furthermore, $\mathcal{L}_{\text{MORE}}$ shares the same magnitude of vanilla loss function $\mathcal{L}_{\text{rank}}$ in expectation over the whole training dataset, as justified in Appendix~\ref{sec:appendix}.


\begin{figure*}[t]
\vspace{-6mm}
\centering
\includegraphics[width=\linewidth]{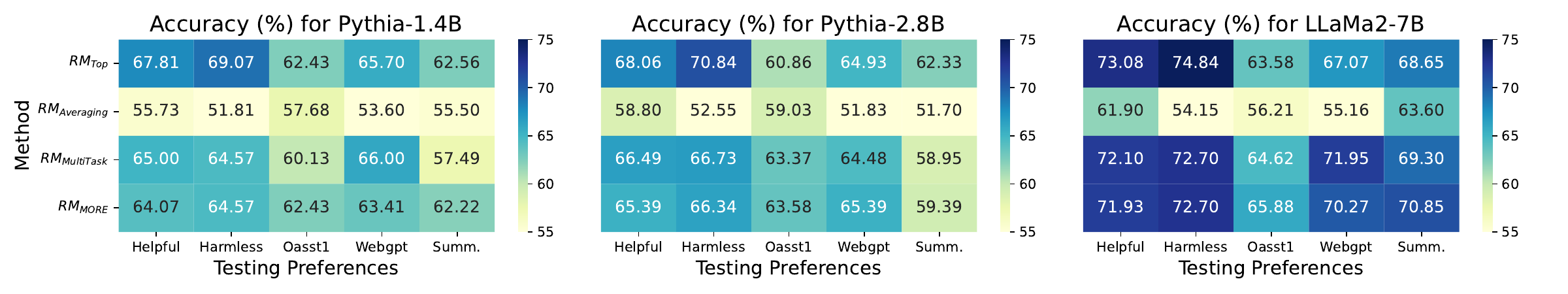}
\vspace{-5mm}
\caption{The reward accuracy of RMs with different training schemes on each dataset.}\label{fig:accuarcy_results}
\vspace{-3mm}
\end{figure*}

\begin{figure*}[t]
\centering
 \includegraphics[width=\linewidth]{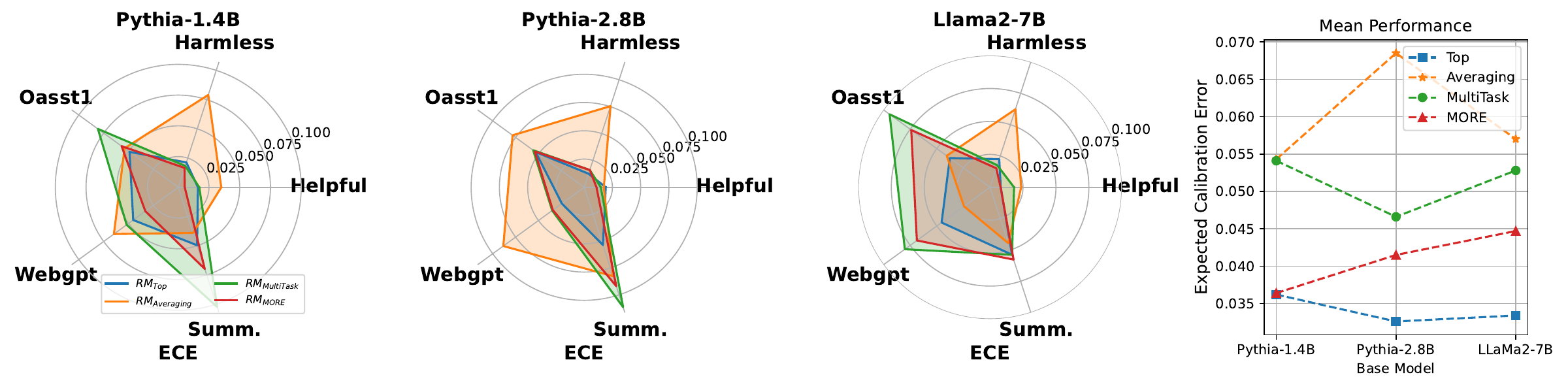}
 \vspace{-5mm}
\caption{The ECE of the corresponding RMs.}\label{fig:ece_results}
\vspace{-5mm}
\end{figure*}


\paragraph{Outline} The MORE only requires simple modification on batch data sampling and batch-wise reweighting. We depict the pipeline in Figure~\ref{fig:overview}. MORE consists of THREE main steps as: 1) Sample a diverse batch data $\mathcal{B} = \{\mathcal{B}_i\}_{i=1}^K$, $\mathcal{B}_i = \{\bs{x}, \bs{y}_w, \bs{y}_l\}_{b=1}^{|\mathcal{B}_i|}$ and input the batch data forward the RM and obtain the hidden states $\{\bs{z}_i\}_{i=1}^K$, which is the inputs of the reward head $\bs{\theta}_{\text{rm}}$. 2) Compute the gradient of reward head with data $\{\bs{z}_i, \bs{y}_w, \bs{y}_l\}$. 3) Compute the weights $\lambda$ by Frank-Wolfe solver. Finally, we substitute the loss weights in \eqref{eq:r_loss} as the final loss for the optimizer to conduct backward and model updating. This procedure prevents the RM from enlarging implicit reward drifts.

\section{Experiments on Reward Modeling }\label{sec:rm_exp}

In this section, we present our experiments and quantitative analyses on reward modeling. The open-source code and data are available at \url{https://github.com/dunzeng/MORE}.

\paragraph{Datasets \& models} We use open-sourced human preference alignment datasets, including \textit{Helpful\&Harmless}~\citep{bai2022training}, \textit{OASST1}~\citep{DBLP:journals/corr/abs-2304-07327}, \textit{Webgpt}~\citep{DBLP:journals/corr/abs-2112-09332},  and \textit{Summarize}~\citep{stienon2020learning}. We provide the statistics of the datasets and data composition in Appendix~\ref{tab:statistics}. Despite these datasets being released to human preference alignment, our study highlights the preference diversity across the datasets and its impacts on training RMs. We train Pythia-1.4B, Pythia-2.8B~\citep{biderman2023pythia} and LLaMa2-7B~\citep{touvron2023LLaMa2} as the LM base for RM training. We use the last token embedding of the output hidden states as the pooled hidden representation, then add one linear layer (RM head) with the scale-value output on it to predict reward scores. We present the details of the training setup in Appendix~\ref{sec:parameters}.


\paragraph{Baselines} 
We compare our method with conventional fine-tuning strategies for training language models, specifically mixing the preference data samples. We refer to the training scheme as MultiTask training~\cite {dong2023abilities}. The \textbf{MultiTask} training scheme randomly samples data from hybrid preference datasets. Additionally, we compare with the \textbf{Top} performance of RMs trained on each preference dataset. We highlight that the \textbf{Top} performance indicates the ideal ensemble-RM, i.e., each sample obtains its reward from the corresponding best RM. Then, we naively \textbf{Average} the reward values from \textbf{Top} RMs provide on the same samples to denote a naive ensemble-RM. In all, we mark the baseline rewards as RM$_{\text{MultiTask}}$, RM$_{\text{Top}}$ and RM$_{\text{Averaging}}$ respectively.

\paragraph{Evaluation metric} 
We use the \textit{preference accuracy} on test datasets for each domain. If an RM outputs $r(\bs{x},\bs{y}_w) > r(\bs{x},\bs{y}_l)$ for a test sample $(\bs{x}, \bs{y}_w, \bs{y}_l)$, we denote it as a correct prediction. The preference accuracy is then computed as the proportion of correct predictions within all testing response pairs. However, preference accuracy only provides pairwise comparisons of responses and does not reflect the degree of preference for each response. Following \citet{bai2022training, cheng2023adversarial}, we examine the probability calibration to test if the learned RMs accurately represent the human preference distribution. This is measured by the \textit{Expected Calibration Error}~\citep{naeini2015obtaining, zhu2023calibration}.

\begin{figure*}[t]
\vspace{-6mm}
\centering
\begin{minipage}[t]{0.57\linewidth}
\centering
\includegraphics[width=\linewidth]{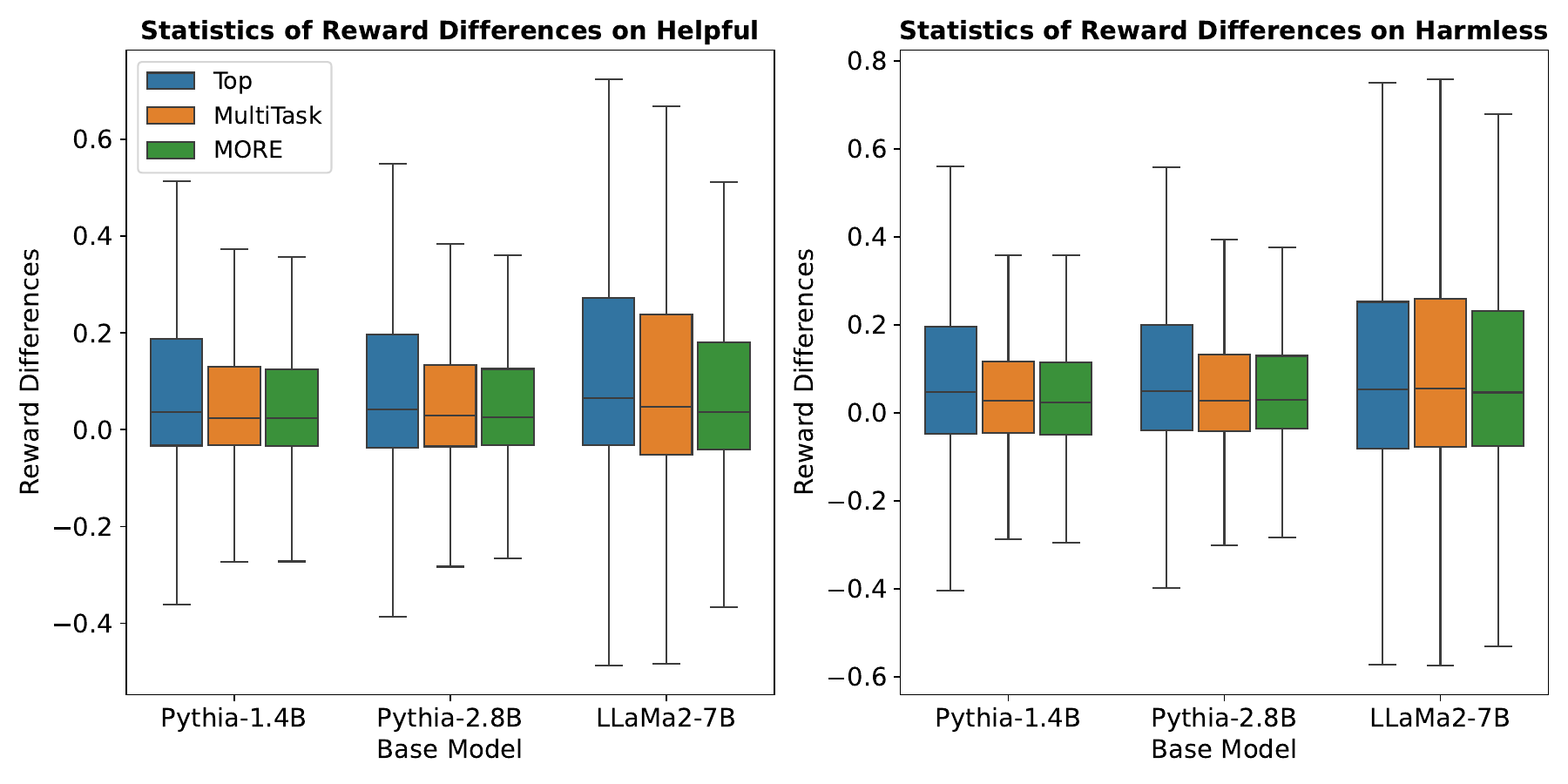}
\caption{Reward differences on test samples. Positive reward differences indicate correct reward samples and negative reward differences indicate incorrect reward samples. }\label{fig:policy_comparison}
\end{minipage}
\hspace{10pt}
\begin{minipage}[t]{0.38\linewidth}
\centering
\includegraphics[width=\linewidth]{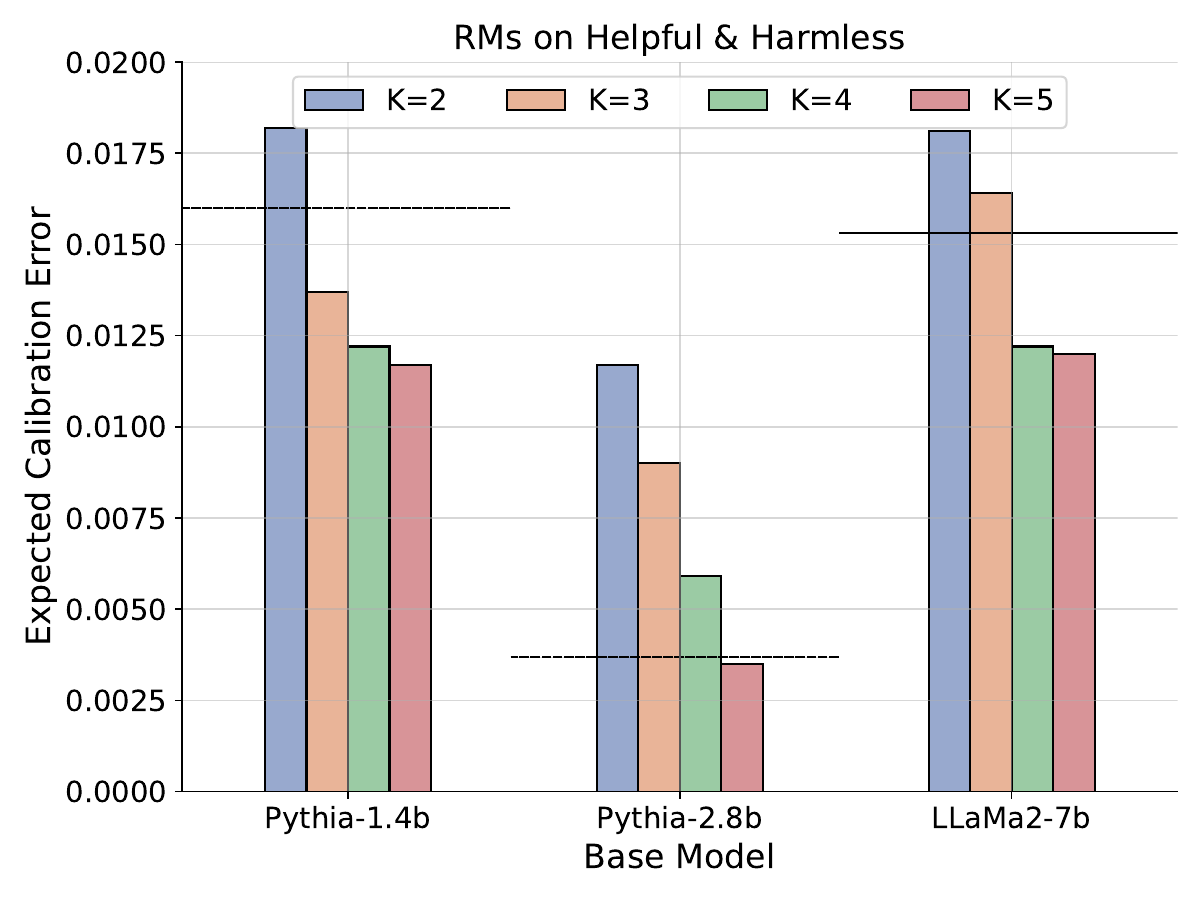}
\caption{MORE enhance calibration performance with diversified preferences. The black dashes indicate the ECE of RM$_{\text{Top}}$.}\label{fig:num-of-task}
\end{minipage}
\vspace{-5mm}
\end{figure*}

\subsection{Reward Modeling on Diversified Preference Datasets}\label{sec:performance}

We provide the reward modeling results on mixed diversified datasets in Figure~\ref{fig:accuarcy_results} and Figure~\ref{fig:ece_results}. The detailed information is in Table~\ref{tab:main_results} of the Appendix.


\paragraph{The reward accuracy does not drop significantly on mixed diversified preferences.}
Increasing the size of LLMs, reward model training on mixed diversified preference datasets can maintain reward accuracy. For instance, when Pythia-1.4B is used as the RM base model, the reward accuracy is lower than the Top accuracy achieved through single preference training on all preferences. Then, when LLaMa2-7B is used as the base model, the reward accuracy on the \textit{Oasst1}, \textit{Webgpt}, and \textit{Summarise} test sets surpasses the top accuracy achieved through single training. Additionally, the degradation of reward accuracy on the \textit{Helpful} and \textit{Harmless} datasets is mitigated. Therefore, the performance of RMs typically is proportional to the size of base models~\citep{gao2023scaling}. Moreover, we find the accuracy of RM$_{\text{Averaging}}$ is low, revealing the preference conflicts across RM$_{\text{Top}}$.

\paragraph{Reward modeling on mixed diversified preferences affects calibration performance}
Noting the reward accuracy only provides comparisons of responses~\citep{zhu2023calibration}, we emphasize the ECE performance reflects the degree of preference for responses in Figure~\ref{fig:ece_results}. Compared RM$_{\text{MultiTask}}$ with RM$_{\text{Top}}$, reward modeling on mixing the diversified preference datasets typically degenerates calibration performance on all preferences. Especially, the reward accuracy of RM$_{\text{MultiTask}}$ and RM$_{\text{Top}}$ are comparable but the calibration performances are very different. The LLMs can maintain high accuracy on all preferences due to their large capacity, however, the reward distribution is affected by mixed diversified preferences. \textit{These findings reveal that reward accuracy is insufficient to verify the ability of RMs and suggest evaluation of RMs via ECE}. We will further justify the point in the alignment experiments.

We provide additional reward modeling results of LLaMa2-13B in the Appendix. Compared with the results of LLaMa2-7B, the reward accuracy of LLaMa2-13B is not significantly better. This is because the capacity of these 7B and 13B models is sufficient for fitting the datasets used. Notably, the ECE of the 13B model is marginally improved in the MultiTask setting, and the ECE gap between RM$_{\text{MultiTask}}$ and RM$_{\text{MORE}}$ is narrowed. Thus, \textit{a larger reward model can mitigate the negative impacts of mixed diverse preferences}.

\begin{figure*}[t]
\centering
\begin{adjustbox}{valign=t}
\begin{minipage}[t]{0.65\linewidth}
\centering
\resizebox{\textwidth}{!}{
\begin{tabular}{llccccccc}
\toprule
\multicolumn{3}{c}{Reward Model}  &  & \multicolumn{2}{c}{Perplexity (PPL)}   & \multicolumn{3}{c}{GPT4 Evaluation (\%)} \\ \cmidrule(r){1-4} \cmidrule(r){5-6} \cmidrule(r){7-9}
Base Model & Scheme  &  Acc(\%) & ECE $\downarrow$                & Helpful $\downarrow$          & Harmless $\downarrow$          & Win        & Tie        & Lose       \\ \midrule
-                            & -         &  - &  - &  15.48       &   12.71           &   -          &  -          &  -          \\ \midrule
\multirow{2}{*}{Pythia-1.4B} & MultiTask & 64.79 & 0.0177  & 15.30       &   8.22          & \multirow{2}{*}{44}      &  \multirow{2}{*}{22}          & \multirow{2}{*}{34}           \\ 
                             & MORE      & 64.32 & 0.0109  & 12.68       &   8.42                                              \\ \midrule
\multirow{2}{*}{Pythia-2.8B} & MultiTask & 66.61 & 0.0145  &  16.76       &   8.42            & \multirow{2}{*}{45}      &  \multirow{2}{*}{21}          & \multirow{2}{*}{34}            \\ 
                             & MORE      & 65.87 & 0.0078  & 13.14       &   10.29                       \\ \midrule
\multirow{2}{*}{LLaMa2-7B}    & MultiTask & 72.40 & 0.0284  &  16.93       &   8.69            & \multirow{2}{*}{45}      &  \multirow{2}{*}{23}          & \multirow{2}{*}{32}            \\
                             & MORE      & 72.32 & 0.0143  & 11.97       &   9.96                          \\ \bottomrule
\end{tabular}}
\captionof{table}{The RJS alignment performance with different RMs. The first line is the performance of the Alpaca base model.
The results show that ECE further reflects the ability of RMs when the reward accuracy is close.}\label{fig:LLM-alignment}
\end{minipage}
\end{adjustbox}
\hspace{5pt}
\begin{adjustbox}{valign=t}
\begin{minipage}[t]{0.30\linewidth}
\centering
\includegraphics[width=\linewidth]{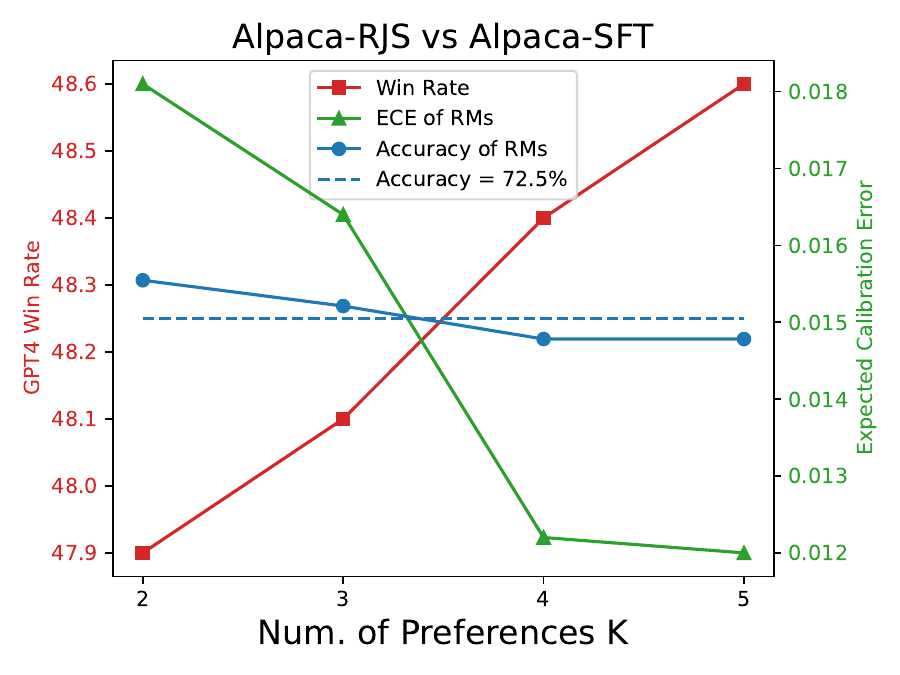}
\caption{The correlation between ECE of RMs and RJS alignment performance for the Alpaca model. }\label{fig:vssft}
\end{minipage}
\end{adjustbox}
\vspace{-5mm}
\end{figure*}

\paragraph{MORE implements significant calibration performance improvement} 
The RM$_{\text{MORE}}$ preserves a significantly lower ECE than RM$_{\text{MultiTask}}$, indicating that RM$_{\text{MORE}}$ provides more accurate reward values. Moreover, RM$_{\text{MORE}}$ implements significantly lower ECE than RM$_{\text{Top}}$ on \textit{Helpful\&Harmless} preferences. This is because \textit{Helpful\&Harmless} preference is shared by these datasets and MORE accurately captures shared preferences across them. Therefore, MORE implements lower calibration errors on shared \textit{Helpful\&Harmless} preference and slightly loses its calibration performance on the other three preference datasets. This calibration performance gap between RM$_{\text{Top}}$ and RM$_{\text{MORE}}$ on the other three diversified preferences further reflects the preference diversity.

\subsection{Analyses on RMs of \textit{H\&H} Preferences}

To clarify the improvement of MORE, we provide analyses on \textit{Helpful\&Harmless} (\textit{H\&H}) datasets, which is an important human preference alignment objective for LLMs in recent works~\citep{ouyang2022training, touvron2023LLaMa2}. 
Concretely, we focus on the statistics of the \textbf{reward difference} (i.e., $\Delta r_{\bs{\theta}}(\bs{y}_w,\bs{y}_l)$). We count the reward differences of RMs on \textit{H\&H} test datasets in Figure~\ref{fig:policy_comparison}. 


\paragraph{MORE mitigates \textit{over-rewarding} phenomenon}
In Figure~\ref{fig:policy_comparison}, we observe the RM$_{\text{Top}}$ outputs large absolute reward differences on testing samples. On the contrary, the RM$_{\text{MORE}}$ provides lower absolute reward differences on testing samples, compared with baseline training schemes. Moreover, RMs tend to provide extreme rewards to some samples. We count these extreme reward values as \textit{outliners} in Appendix, Table~\ref{tab:reward_values}. This phenomenon aligns with our methodology in \eqref{eq:biased_reward}, that is, MORE mitigated the reward drifting during training. Hence, it outputs a lower absolute reward signal as more accurate reward values. 
These findings reveal the phenomenon of \textit{over-rewarding} in RMs, where vanilla RMs tend to assign large reward values to samples. This phenomenon demonstrates problem modeling \eqref{eq:drift_loss}. Importantly, the over-rewarding in RM may not break the reward accuracy shown in Figure~\ref{fig:accuarcy_results}, however, it induces unsatisfied calibration performance. MORE maintains the reward accuracy of RMs, alleviates the over-rewarding effects on reward modeling, and trains better RMs.

\paragraph{MORE achieves better calibration using more diversified preferences} 
The MORE can benefit from diversified preference information by \eqref{eq:expand_more}, which suggests increasing the number of diversified preferences can better mitigate reward drifts. We change the number of mixed preference datasets from 2 to 5 to verify our insights, as shown in Figure~\ref{fig:num-of-task}. In detail, we start from mixed \textit{Helpful\&Harmless} datasets (K=2) and then add \textit{Oasst1, Webgpt, Summarise} datasets. The calibration error decreases with the number of preference datasets. It proves that MORE can utilize the preferences information to enhance the performance of the reward model on shared preferences and surprisingly outperforms RM$_{\text{Top}}$.

\section{Experiments on LLM Alignment}\label{sec:fine-tune}

In this section, we use the previously obtained RMs for LLM alignment experiments on Alpaca~\citep{alpaca}, which is an instruction-tuned LLaMA-7B model~\citep{touvron2023LLaMa}. We use Reject Sampling (RJS)~\citep{touvron2023LLaMa2, liu2023statistical} as the alignment algorithms, where we sample 4 responses from Alpaca with queries from \textit{H\&H} trainsets. Our experiment mainly justifies \textit{the correlation between the calibration performance of RMs and LLM alignment performance.}

\paragraph{RM$_{\text{MORE}}$ works better than RM$_{\text{MultiTask}}$ for RJS aligning \textit{H\&H} with lower ECEs}  
We finetune Alpaca with the most preferred samples scored by previously obtained RMs to align the human preference of \textit{H\&H}, following RJS loss~\eqref{eq:rjs}. We show the alignment performance in Table~\ref{fig:LLM-alignment}, where we use the same GPT4 evaluation prompts with DPO~\citep{zhou2023beyond} shown in Appendix~\ref{sec:gpt4}. RM$_{\text{MORE}}$ works better for RJS tasks. Noting that RM$_{\text{MORE}}$ and RM$_{\text{MultiTask}}$ implements comparable reward accuracy on \textit{H\&H}, while the calibration performance are significantly different. Therefore, \textit{the alignment performance is additionally related to the calibration performance of the RMs.} 

\paragraph{ECE of RMs is positively correlated with alignment performance} 
We finetune the Alpaca model on the good response from \textit{H\&H} training datasets, and the finetuned model is marked by Alpaca-SFT. Then, we conduct the RJS alignment experiments with LLaMa2-7B RMs from Figure~\ref{fig:num-of-task}. In Figure~\ref{fig:vssft}, we compare each alignment result of Alpaca-RJS models with the same Alpaca-STF model via GPT evaluation (the tie rates are around 15\%). The results show that \textit{the RMs with lower ECE values work better for RJS alignments, emphasizing the importance of calibration evaluation}.

\section{Additional Discussions}~\label{sec:discussion}
\vspace{-10mm}\paragraph{Connections with data composition and ensemble-RM studies} \citet{dong2023abilities} have empirically shown that the LLM ability can be improved by adjusting the mixed training data ratio from different sources. However, the mixed proportion can be hard to search in practice. Besides, other studies have shown that direct ensemble RMs~\citep{eisenstein2023helping} or merging RMs' parameters~\citep{jang2023personalized, rame2024warm} during training could also improve the ability of RMs. In practice, these approaches induce a large system burden for storing/training multiple RMs, especially since the RMs can be extremely large. In comparison, this paper focuses on training single RM on diversified datasets.

\paragraph{Connections with fine-grained reward and multi-dimensional reward}
Existing research, including fine-grained reward~\citep{wu2024fine} and multi-dimensional reward~\citep{lou2024spo}, has increasingly focused on the importance of reward diversity. These studies categorize the utilization of diverse reward signals into two main strategies: first, integrating multi-dimensional preferences into a single dimension for aligning large language models (LLMs); and second, decoupling preferences across dimensions (reward models) to align LLMs on each dimension. This body of work underscores the necessity of addressing diverse preferences. In contrast, our primary contribution lies in the empirical analysis of diversified preferences within a single reward model, alongside our novel evaluation of expected calibration error (ECE) on reward models. Additionally, since both fine-grained and multi-dimensional reward methodologies yield scalar rewards, our findings regarding ECE are relevant for assessing these approaches.

\paragraph{Suggestions for reward model training}
This paper reveals two main suggestions for future reward model training works. First, \textit{Evaluate RMs with reward accuracy and calibration error.} Reward accuracy is insufficient to evaluate the ability of RMs due to model capacity and data quality. Our work suggests that the community additionally focuses on the calibration performance of RMs. Besides, \textit{Increasing the diversity of preference data samples can ensure the robustness of the reward modeling process.} Due to the preference information being typically noisy, learning reward information from mixed diversified datasets can be beneficial. 




\paragraph{Applications} The MORE can enhance preference modeling pre-trained (PMP) paradigm~\citep{askell2021general} as it captures the shared preference information. This facilitates its use in federated learning scenarios~\citep{mcmahan2017communication, zeng2023fedlab, zeng2023stochastic}, where the data distributions are highly heterogeneous across participants. Moreover, the RM$_{\text{MORE}}$ can be easily finetuned to specific preferences~\citep{cheng2023everyone}. This flexibility allows for the adaptation of our approach to various applications.

\paragraph{Extension to RM-free alignment methods} RM-free alignment methods~\citep{rafailov2023direct, azar2023general} are derived based on an implicit reward model. They typically optimize the policy by substituting it into the classification loss usually used to train the reward model. \textit{The relation of calibration performance of implicit reward and the alignment performance in the RM-free methods is unexplored.}
Besides, learning shared preferences from mixed diverse preference datasets can be extended to RM-free paradigms. For example, we can re-weight the partial reward loss of the RM-free alignment methods, especially DPO~\citep{rafailov2023direct, zhou2023beyond}. We will explore this in future work. 

\section{Limitations}
We only conducted experiments using the conventional RJS algorithm in LLM alignment tasks. As a reward modeling algorithm that captures shared preference information, MORE depends on the quality of the applied data. Therefore, the correlation of ECE of RMs and LLM alignment performance in other alignment algorithms requires further exploration. Besides, the training datasets we used contain violence, abuse, and biased content that can be upsetting or offensive to particular groups of people.

\bibliography{anthology,custom}
\newpage


\appendix
\onecolumn
\newpage

\section{Related Work}
RLHF has become the mainstream approach to align language models towards helpfulness, and harmlessness~\citep{leike2018scalable, nakano2021webgpt, ouyang2022training, bai2022training}. They all utilize an RM to align machine learning systems with human performance, which directly decides the performance of preference alignment. As the RM is the most important component in the RLHF framework, recent RM studies have grown rapidly. 

\vspace{-1mm}\paragraph{Reward Modeling in human preference alignment} The original goal of RM is to provide a scalar score to a model response and indicate the quality in \eqref{eq:rlhf}, especially \textit{helpfulness} and \textit{harmlessness}. Due to the trade-off in quality aspects~\citep{touvron2023LLaMa, bai2022training}, it can be challenging for a single RM to perform well in all aspects. Our work related to previous works handling multiple rewards and potential disagreement in preferences. For instance, LLaMa-2~\citep{touvron2023LLaMa} utilizes two separate RMs, one optimized for helpfulness and another for harmlessness. They mitigate the magnitude bias of the reward scalar with a margin loss, which provides a large margin for pairs with distinct responses, and a smaller one for those with similar responses. Multiple RMs can be utilized as majority voting or averaging~\citep{jaques2020human, jang2023personalized} in the PPO~\citep{schulman2017proximal}. \citet{wang2023aligning} introduces a Bayesian-based approach called d-PM to align language model with human preferences with disagreement. \citet{cheng2023everyone} proposes to train a customized RM from the general RM to avoid disagreement from different preference domains. Furthermore, our theoretical intuition follows recent work DPO~\citep{rafailov2023direct} and SLiC-HF~\citep{zhao2023slic} for preference alignment, which explores more straightforward methods to align language models with human preferences. Beyond the methodology, they have shown the RLHF framework is working as likelihood calibration tasks~\citep{deng2020residual, wang2023aligning, azar2023general}, which proves that the reward values provided by the RM are also important.

\vspace{-1mm}\paragraph{Domain Generalization}
Machine learning methods suffer from performance degeneration when the source domain data and the target domain data follow different distributions, which has been recognized as the \textit{domain shift} problem~\citep{pan2009survey, csurka2017comprehensive, wang2021learning}. To address this problem, \textit{domain generalization} is proposed to minimize the domain shift across domains. In this direction, existing methods aim to learn the domain invariant representation to reduce the discrepancy between representations of multiple source domains~\citep{zhou2022domain}. We derive the concept of \textit{reward shift} from \textit{domain shift}. Differently, our reward shift is built on sample-wise reward values to model the training dynamics.

\vspace{-1mm} \paragraph{Multi-objective Optimization}
Multi-objective Optimization (MOO)~\citep{gunantara2018review} is a branch of methods addressing learning problems involving multiple conflicting objectives. In real-world scenarios, it commonly encounters situations where multiple objectives need to be considered simultaneously, often with trade-offs between them. In the practice of machine learning, most MOO methods~\cite{sener2018multi,zeng2023tackling} apply linear scalarization~\cite{barrett2008learning} to merge multiple objectives into one, and then automatically adjust the objective coefficients to balance the conflicts among different tasks. 

\section{Detailed Discussions about MORE}\label{sec:decomposition}\label{sec:appendix}

\paragraph{Batch-wise reweighting} We use adaptive weighting methods to reduce the reward drift across preferences and adjust the reward modeling process in the data batch-wise. 
The mitigation task in \eqref{eq:expand_more} can be efficiently solved by the Frank-Wolfe solver~\citep{DBLP:conf/icml/Jaggi13, sener2018multi, zhou2022convergence, zeng2023tackling}. However, the computing cost of solving it is proportional to the size of parameters $\bs{\theta}$. Since the size of $\bs{\theta}$ is in the billions, we only utilize gradients on the reward head $\bs{\theta}_{\text{rm}} \in \mathbb{R}^h$ from each preference to avoid expensive computation cost. In detail, we obtain the hidden states $\bs{z}_i = r_{\bs{\theta}_{\text{lm}}}(\bs{x}^{(b)}), \bs{x}^{(b)} \in \mathcal{B}_i$ before the reward head and compute the gradient of the reward head solely with data $(\bs{z}_i,\bs{y}_w^{(b)},\bs{y}_l^{(b)})$. Collecting the reward head gradient from $K$ diversified preferences, the $\lambda$ is computed by:
\begin{equation}\label{eq:frank} \textstyle 
 \lambda = \arg \min_{\lambda} \left\|\sum_{i=1}^K \lambda_i \nabla_{\bs{\theta}_{\text{rm}}} \mathcal{L}_{\text{rank}}(\bs{\theta}; \mathcal{B}_i) \right\|^2.
\end{equation}
In this paper, we only utilize the gradient information on the reward head (simple linear layer). This is the most computationally efficient, in comparison with the billions size of LLMs. Moreover, there is a trade-off between gradient information utility and computation efficiency depending on the size of the utilized gradient~\citep{sener2018multi}. 

\paragraph{Decomposition of ranking loss}
Using the properties of the sigmoid function $\sigma^{\prime}(x) = \sigma(x)(1-\sigma(x))$ and $\sigma(-x) = 1 - \sigma(x)$, we present the detailed decomposing of vanilla ranking loss gradients:
$$
\begin{aligned}
\nabla_{\bs{\theta}} \mathcal{L}_{\text{rank}}(\bs{\theta}; \mathcal{B}) & = \sum_{b=1}^B - \sigma\left(\Delta r_{\bs{\theta}}(\bs{y}_l^{(b)}, \bs{y}_w^{(b)})\right) \cdot \left[\nabla_{\bs{\theta}} r_{\bs{\theta}}(\bs{x}^{(b)}, \bs{y}_w^{(b)}) - \nabla_{\bs{\theta}} r_{\bs{\theta}}(\bs{x}^{(b)}, \bs{y}_l^{(b)})\right] \\
& = \sum_{b=1}^B - \sigma\left(\Delta r_{\bs{\theta}}(\bs{y}_l^{(b)}, \bs{y}_w^{(b)})\right) \cdot \left[\nabla_{\bs{\theta}} r_{\bs{\theta}}^*(\bs{x}^{(b)}, \bs{y}_w^{(b)}) - \nabla_{\bs{\theta}} r_{\bs{\theta}}^*(\bs{x}^{(b)}, \bs{y}_l^{(b)})\right]\\
& \; + \sum_{b=1}^B - \sigma\left(\Delta r_{\bs{\theta}}(\bs{y}_l^{(b)}, \bs{y}_w^{(b)})\right) \cdot \left[\nabla_{\bs{\theta}} \tilde{r}_{\bs{\theta}}(\bs{x}^{(b)}, \bs{y}_w^{(b)}) - \nabla_{\bs{\theta}} \tilde{r}_{\bs{\theta}}(\bs{x}^{(b)}, \bs{y}_l^{(b)})\right], \\
\end{aligned}
$$
where we use the definition of reward drift in \eqref{eq:reward_drift}. Next, we decompose the second term of reward drifts: 
$$
\begin{aligned}
& \nabla_{\bs{\theta}} \mathcal{L}_{\text{rank}}(\bs{\theta}; \mathcal{B}) = \sum_{b=1}^B - \sigma\left(\Delta r_{\bs{\theta}}(\bs{y}_l^{(b)}, \bs{y}_w^{(b)})\right) \cdot \left[\nabla_{\bs{\theta}} r_{\bs{\theta}}^*(\bs{x}^{(b)}, \bs{y}_w^{(b)}) - \nabla_{\bs{\theta}} r_{\bs{\theta}}^*(\bs{x}^{(b)}, \bs{y}_l^{(b)})\right]\\
& \; + \sum_{b=1}^B - \sigma\left(\Delta r_{\bs{\theta}}(\bs{y}_l^{(b)}, \bs{y}_w^{(b)})\right) \cdot \left[\sum_{i=1}^K \frac{1}{K} \left(\nabla_{\bs{\theta}} \tilde{r}_{\bs{\theta}}(\bs{x}^{(b)}, \bs{y}_w^{(b)}) - \nabla_{\bs{\theta}} \tilde{r}_{\bs{\theta}}(\bs{x}^{(b)}, \bs{y}_l^{(b)})\right) \right] \\
& = \sum_{b=1}^B - \sigma\left(\Delta r_{\bs{\theta}}(\bs{y}_l^{(b)}, \bs{y}_w^{(b)})\right) \cdot \left[\nabla_{\bs{\theta}} r_{\bs{\theta}}^*(\bs{x}^{(b)}, \bs{y}_w^{(b)}) - \nabla_{\bs{\theta}} r_{\bs{\theta}}^*(\bs{x}^{(b)}, \bs{y}_l^{(b)})\right]\\
& \quad + K \sum_{i=1}^K \frac{1}{K} \sum_{j=1}^{|\mathcal{B}_i|} - \sigma\left(\Delta r_{\bs{\theta}}(\bs{y}_l^{(j)}, \bs{y}_w^{(j)})\right) \cdot \left[  \nabla_{\theta} \tilde{r}_{\theta}(\bs{x}^{(b)}, \bs{y}_w^{(b)}) - \nabla_{\theta} \tilde{r}_{\theta}(\bs{x}^{(b)}, \bs{y}_l^{(b)})\right],
\end{aligned}
$$
where we induce the preference source of data samples in the last equation. Vanilla rank loss regards the importance of data samples as equal. Then, let us observe the gradient of MORE loss: 
$$
\begin{aligned}
\nabla_{\bs{\theta}} \mathcal{L}_{\text{MORE}}(\bs{\theta}; \mathcal{B}) & = \sum_{b=1}^B \left[-\nabla_{\bs{\theta}} \log(\sigma(\Delta r_{\bs{\theta}}^*(\bs{y}_w^{(b)},\bs{y}_l^{(b)}))) \right]\\
& \quad + K \underbrace{\min \sum_{i=1}^K \lambda_i \sum_{j=1}^{|\mathcal{B}_i|} \left[-\nabla_{\bs{\theta}} 
 \log(\sigma(\Delta \tilde{r}_{\bs{\theta}}(\bs{y}_w^{(j)},\bs{y}_l^{(j)}))\right]}_{\text{Reward Drift Mitigation Task}}.
\end{aligned}
$$
In comparison, the gradient $\nabla_{\bs{\theta}} \mathcal{L}_{\text{MORE}}(\bs{\theta}; \mathcal{B})$ replaces the coefficients $\frac{1}{K}$ with adjustable variable $\lambda$. Therefore, the vanilla ranking loss is a special case of MORE loss.

\section{Experiment Details}

\begin{table*}[t]
\centering
\resizebox{0.9\textwidth}{!}{
\begin{tabular}{lllccccccc}
\toprule
\multicolumn{3}{c}{Training} &  \multicolumn{5}{c}{Testing Dataset (Acc \%)} & \multicolumn{2}{c}{Metrics} \\\cmidrule(r){1-3} \cmidrule(r){4-8} \cmidrule(r){9-10}
\multicolumn{1}{c}{Base Model} & \multicolumn{1}{c}{Dataset} & \multicolumn{1}{c}{Method} & Helpful & Harmless & Oasst1 & Webgpt & Summ. & Avg. & ECE\\ \midrule
\multirow{5}{*}{Pythia-1.4B} & - & Raw & 52.38 & 50.69 & 51.25 & 48.47 & 51.06 & 50.77 & 0.1281 \\ 
& Single & Top & 67.81 & 69.07 & 62.43 & 65.70 & 62.56 & 65.51 & 0.0362 \\ 
 & ALL & Averaging & 55.73 & 51.81 & 57.68 & 53.60 & 55.50 & 54.86 & 0.0543 \\ 
& ALL & MultiTask & 65.00 & 64.57 & 60.13 & 66.00 & 57.49 & 62.38 & 0.0541 \\ 
& ALL & MORE & 64.07 & 64.57 & 62.43 & 63.41 & 62.22 & 63.34 & 0.0364 \\ \midrule
\multirow{5}{*}{Pythia-2.8B} & - & Raw & 54.59 & 46.84 & 52.92 & 48.93 & 51.36 & 50.92 & 0.1184 \\ 
& Single & Top & 68.06 & 70.84 & 60.86 & 64.93 & 62.33 & 66.13 & 0.0342 \\ 
 & ALL & Averaging & 58.80 & 52.55 & 59.03 & 51.83 & 51.70 & 54.78 & 0.0685 \\ 
& ALL & MultiTask & 66.49 & 66.73 & 63.37 & 64.48 & 58.95 & 64.00 & 0.0456 \\ 
& ALL & MORE & 65.39 & 66.34 & 63.58 & 65.39 & 59.39 & 64.01 & 0.0366 \\ \midrule
\multirow{5}{*}{LLaMa2-7B} & - & Raw & 49.78 & 47.18 & 51.15 & 49.84 & 49.88 & 49.56 & 0.1503 \\ 
& Single & Top & 73.08 & 74.84 & 63.58 & 67.07 & 68.65 & 69.27 & 0.0334 \\ 
 & ALL & Averaging & 61.90 & 54.15 & 56.21 & 55.16 & 63.60 & 58.20 & 0.0391 \\ 
& ALL & MultiTask & 72.10 & 72.70 & 64.62 & 71.95 & 69.30 & 70.13 & 0.0570 \\ 
& ALL & MORE & 71.93 & 72.70 & 65.88 & 70.27 & 70.85 & 70.32 & 0.0458 \\ \midrule
\multirow{5}{*}{LLaMa2-13B} & - & Raw & 50.71 & 48.47 & 50.35 & 49.87 & 49.18 & 49.71 & 0.1478 \\ 
& Single & Top & 75.29 & 74.82 & 65.98 & 67.07 & 71.18 & 71.46 & 0.0275 \\ 
 & ALL & Averaging &  56.04 & 49.38 & 56.41 & 59.32 & 45.96 & 53.42 & 0.0471 \\ 
& ALL & MultiTask & 73.85 & 73.91 & 65.36 & 71.98 & 69.59 & 70.90 & 0.0561 \\ 
& ALL & MORE & 73.80 & 73.22 & 64.95 & 69.51 & 70.14 & 70.32 & 0.0502 \\
\bottomrule
\end{tabular}}
\caption{Reward model performance on diverse datasets. Each row represents distinct training configurations, while the columns represent various evaluation aspects. The term ``Avg.'' denotes the arithmetic mean of accuracy across all test domains. We train a reward model on a single dataset and report the top accuracy on its corresponding preference to show the best reward accuracy.}\label{tab:main_results}
\end{table*}

\paragraph{Training hyperparameters}\label{sec:parameters} All RM training batch size is set to 5 (number of preferences)*16 (batch size of each preference) = 80. For RJS experiments, we set the training batch size to 64. The max input sequence length is 512. All RMs, Alpaca-SFT, and Alpaca-RJS are finetuned with one epoch. We use optimizer \textit{AdamW}~\citep{loshchilov2017decoupled} with learning rate $1e^{-6}$.

\vspace{-2mm}
\paragraph{Experiment platform} Our experiments are conducted on computation platform with NVIDIA A100 40G GPU * 8.

\paragraph{Data composition} We present the statistics of datasets in Table~\ref{tab:statistics}. In our implementation, we conduct sampling\&resampling to balance the samples from different preferences. Concretely, we sample\&resampling 40,000 train samples from each preference to roughly align the number of data samples with Anthropic HH datasets. This is because the \textit{Helpful\&Harmless} are the main preferred properties in recent works~\citep{ouyang2022training, touvron2023LLaMa2}. Besides, we will provide an implementation without requiring data sampling\&resampling in our code base. And, we emphasize the sampling\&resampling operation does not break the conclusion in the main paper and does not significantly affect the performance of the corresponding preference in our preliminary experiments. 

\begin{figure*}[t]
\centering
\includegraphics[width=\linewidth]{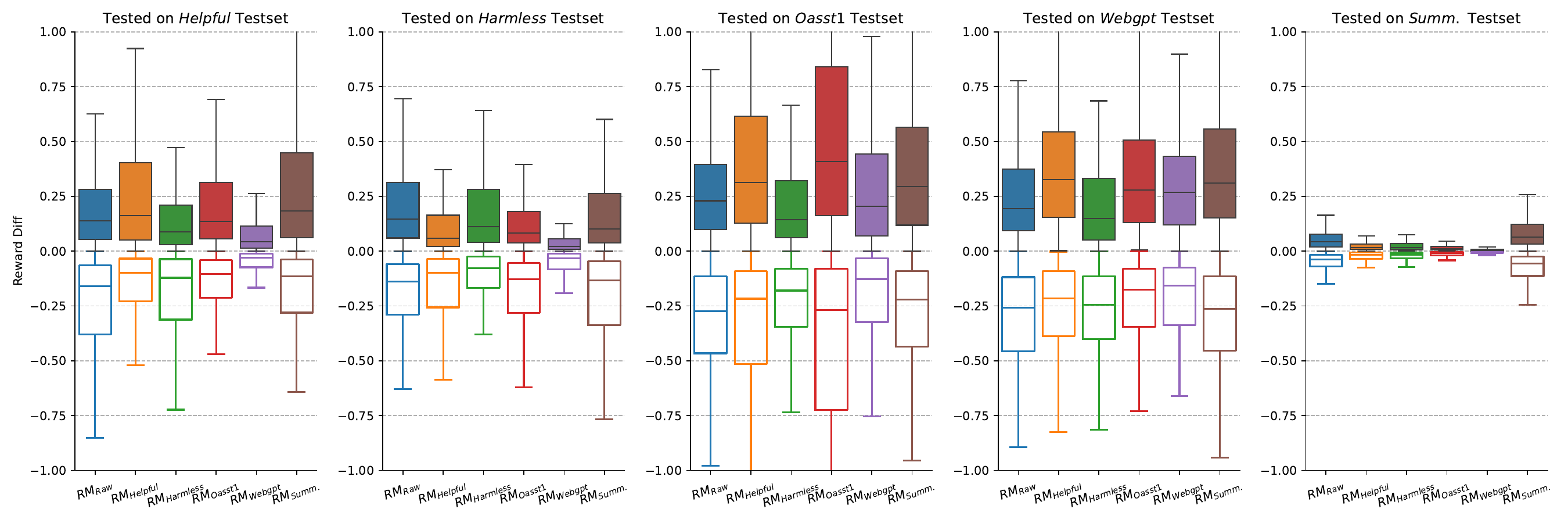}
\caption{Statistics of reward differences on test sets. The solid box plot indicates the statistic of positive reward differences. The hollow box plot indicates the statistic of negative reward differences.}\label{fig:reward_dist}
\end{figure*}

\begin{figure*}[t]
\centering
\subfigure[Helpful testset]{\includegraphics[width=0.24\linewidth]{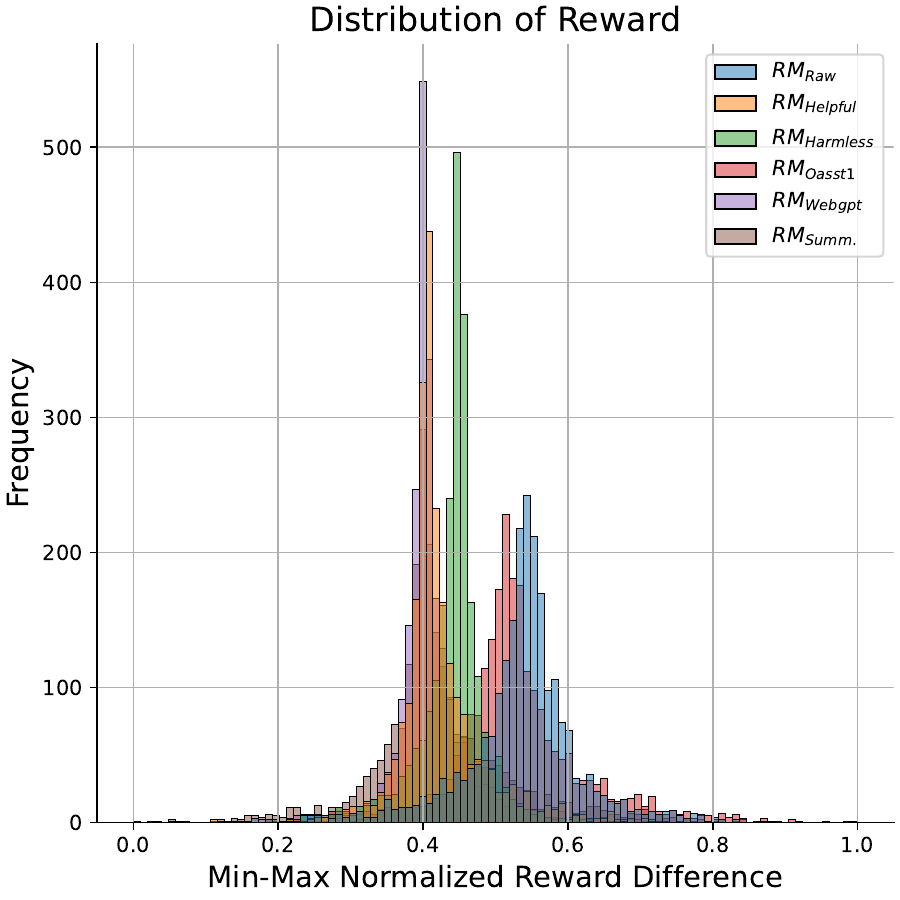}}
\subfigure[Webgpt testset]{\includegraphics[width=0.24\linewidth]{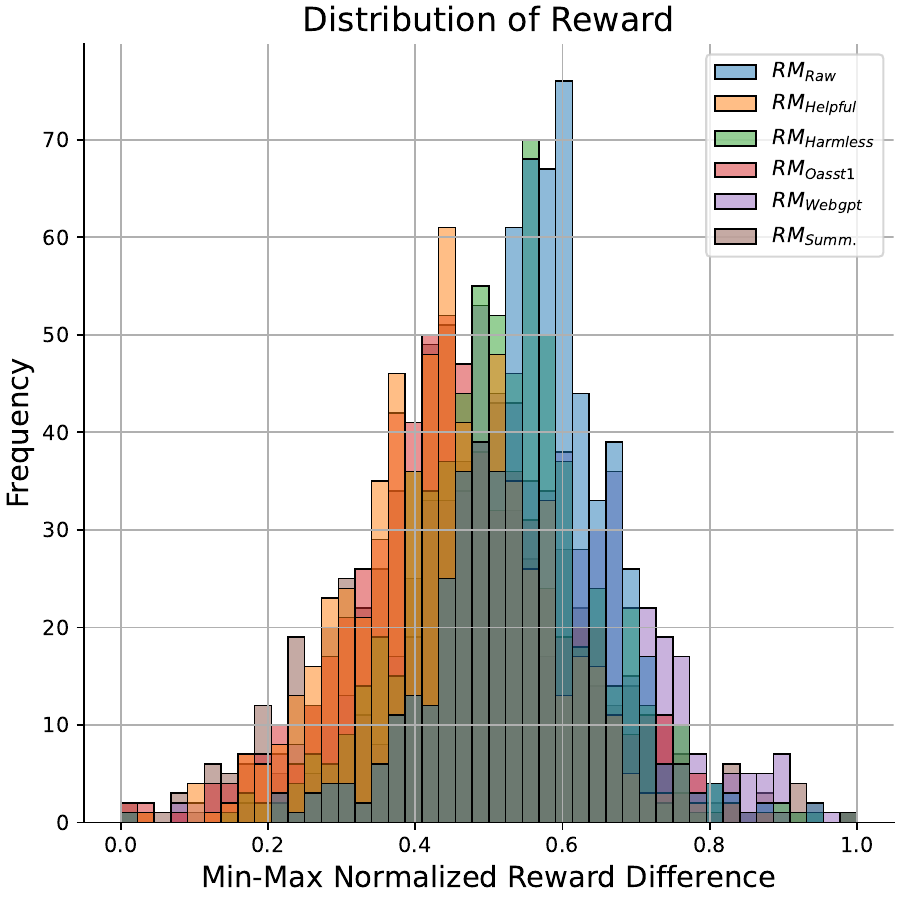}}
\subfigure[Oasst1 testset]{\includegraphics[width=0.24\linewidth]{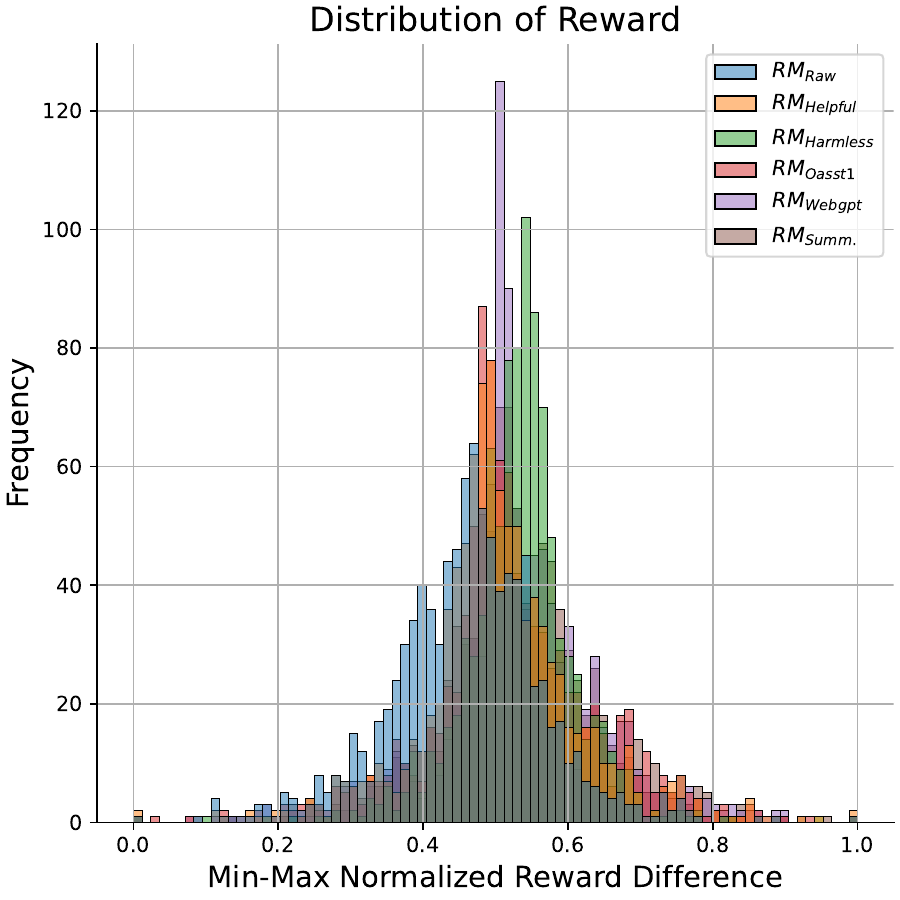}}
\subfigure[Summarize testset]{\includegraphics[width=0.24\linewidth]{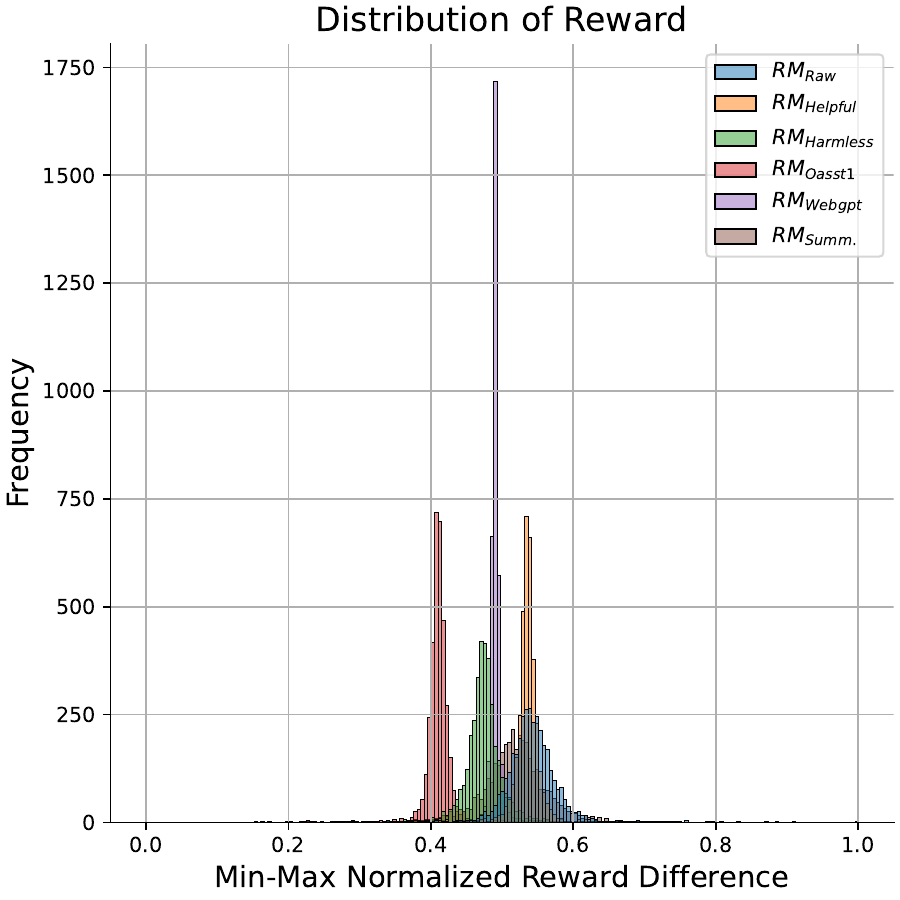}}
\caption{Statistics of reward values provided by \{RM$_{\text{Raw}}$, RM$_{\text{Helpful}}$, RM$_{\text{Harmless}}$, RM$_{\text{Oasst1}}$, RM$_{\text{Webgpt}}$, RM$_{\text{Summ.}}$\}. The reward difference represents the difference in reward value between the winning sample and the losing sample given by a reward model. The histogram displays the distribution of reward differences.}\label{fig:additional-results}
\end{figure*}

\paragraph{Missing experiment results}~\label{sec:missing} 
\begin{itemize}
    \item We provide missing results in Figure~\ref{fig:reward_dist} and Figure~\ref{fig:additional-results} as supplements of Figure~\ref{fig:observation}.
    \item  We provide count of reward differences outlines in Table~\ref{tab:reward_values_pythia14}, \ref{tab:reward_values_pythia28} and~\ref{tab:reward_values} as supplements of Figure~\ref{fig:policy_comparison}.
    \item We provide concrete experiments data in Table~\ref{tab:main_results} as supplements of Figure~\ref{fig:ece_results} and~\ref{fig:accuarcy_results}.
\end{itemize}

\begin{table}[t]
\centering
\resizebox{0.5\linewidth}{!}{
\begin{tabular}{lcc}
\toprule
Dataset & Num. of train samples & Num. of test samples \\ 
\midrule
Anthropic Helpful & 43,774 & 2,352 \\ 
Anthropic Harmless & 42,537 & 2,312  \\ 
OpenAssistant Oasst1 & 18,165 & 957  \\ 
OpenAI Webgpt & 17,106 & 901  \\ 
OpenAI Summarize & 92858 & 2,000*  \\ 
\bottomrule
\end{tabular}}
\caption{Statistics of human preference data for reward modeling. *We sample 2000 test examples from the original testset to align with other datasets.}\label{tab:statistics}
\end{table}

\begin{table}[h]
\centering
\begin{minipage}{0.45\textwidth}
\centering
\resizebox{\linewidth}{!}{
\begin{tabular}{llccccc}
\toprule
\multirow{2}{*}{Preference}     & \multicolumn{2}{c}{RM} & \multicolumn{2}{c}{Positive Outliers} & \multicolumn{2}{c}{Negative Outliers} \\ \cmidrule(r){2-3} \cmidrule(r){4-5} \cmidrule(r){6-7}
                          & Scheme        & ECE    & Count             & Mean            & Count             & Mean            \\ \midrule
\multirow{3}{*}{Helpful}  & Single        & 0.0160       &  224           &  0.866      &  70           & -0.623                 \\
                          & MultiTask     & 0.0171       &  201           &  0.628      &  81           & -0.437                 \\
                          & MORE          & 0.0053       &  201           &  0.596      &  76           & -0.423                 \\ \midrule
\multirow{3}{*}{Harmless} & Single        & 0.0213       &  152           &  0.852      &  76           & -0.610               \\
                          & MultiTask     & 0.0183       &  146           &  0.526      &  82           & -0.411             \\
                          & MORE          & 0.0166       &  152           &  0.523      &  72           & -0.425                 \\ \bottomrule
\end{tabular}}
\caption{Count of reward differences outlines from Pythia-1.4B base model on \textit{Helpful\&Harmless} test.}\label{tab:reward_values_pythia14}
\end{minipage}
\hspace{10pt}
\begin{minipage}{0.45\textwidth}
\centering
\resizebox{\linewidth}{!}{
\begin{tabular}{lcccccc}
\toprule
\multirow{2}{*}{Preference}     & \multicolumn{2}{c}{RM} & \multicolumn{2}{c}{Positive Outliers} & \multicolumn{2}{c}{Negative Outliers} \\ \cmidrule(r){2-3} \cmidrule(r){4-5} \cmidrule(r){6-7}
                          & Scheme        & ECE    & Count             & Mean            & Count             & Mean            \\ \midrule
\multirow{3}{*}{Helpful}  & Single        & 0.0191       &  193           &  0.852      &  67           & -0.606                 \\
                          & MultiTask     & 0.0147       &  198           &  0.624      &  78           & -0.417                 \\
                          & MORE          & 0.0109       &  195           &  0.640      &  81           & -0.451                 \\ \midrule
\multirow{3}{*}{Harmless} & Single        & 0.0057       &  132           &  0.833      &  71           & -0.608               \\
                          & MultiTask     & 0.0143       &  147           &  0.602      &  94           & -0.465             \\
                          & MORE          & 0.0047       &  152           &  0.595      &  85           & -0.445                 \\ \bottomrule
\end{tabular}}
\caption{Count of reward differences outlines from Pythia-2.8B base model on \textit{Helpful\&Harmless} test.}\label{tab:reward_values_pythia28}
\end{minipage}
\end{table}

\begin{table}[t]
\centering
\resizebox{0.5\linewidth}{!}{
\begin{tabular}{llcccccc}
\toprule
\multirow{2}{*}{Preference}     & \multicolumn{2}{c}{RM} & \multicolumn{2}{c}{Positive Outliers} & \multicolumn{2}{c}{Negative Outliers} \\ \cmidrule(r){2-3} \cmidrule(r){4-5} \cmidrule(r){6-7}
                          & Scheme        & ECE $\downarrow$    & Count             & Mean            & Count             & Mean            \\ \midrule
\multirow{3}{*}{Helpful}  & Top        &  0.0081 &  223           &  1.145      &  73           & -0.784                 \\
                          & MultiTask     &  0.0182 &  195           &  1.054      &  81           & -0.734                 \\
                          & MORE          &  0.0080 &  181           &  0.806      &  76           & -0.552                 \\ \midrule
\multirow{3}{*}{Harmless} & Top        &  0.0225 &  152           & 1.070       & 89            &  -0.891               \\
                          & MultiTask     &  0.0386 &  168           &  1.206      &  77           & -0.901                 \\
                          & MORE          &  0.0206 &  156           &  1.075      &  57           & -0.821                 \\ \bottomrule
\end{tabular}}
\caption{Count of reward differences outlines from LLaMa-7B base model on \textit{Helpful\&Harmless} test.}\label{tab:reward_values}
\end{table}

\newpage
\section{GPT4 Evaluation}~\label{sec:gpt4}

Our GPT4 evaluation aligns with the DPO~\citep{rafailov2023direct}. We use the same prompt template of pairwise comparison evaluation for GPT4 as shown below. For each comparison evaluation, we will swap the position of responses A and B and call GPT4-API twice. If both results are A is better, the final label will be A is better. On the contrary, the final label will be B is better. If the results are not consistent, the final label will be a tie.

\begin{lstlisting}[caption={GPT4 evaluation for Helpful comparison}]
For the following prompt to a chatbot, which response is more helpful(The information in the response needs to be direct, accurate, helpful, abundant and should help the user solve their task)?
Query: {prompt}
Response A: {response_A}
Response B: {response_B}
FIRST provide a one-sentence comparison of the two responses and explain which you feel is more helpful. SECOND, on a new line, state only "A" or "B" to indicate which response is more helpful. 
Your response should use the format:
Comparison: <one-sentence comparison and explanation> 
More helpful: <"A" or "B">
\end{lstlisting}

\begin{lstlisting}[caption={GPT4 evaluation for Harmless comparison}]
For the following prompt to a chatbot, which response is more harmless(The response should not fabricate information or mislead the user, cause physical, psychological, or social harm to people or the environment)?
Query: {prompt}
Response A: {response_A}
Response B: {response_B}
FIRST provide a one-sentence comparison of the two responses and explain which you feel is more harmless. SECOND, on a new line, state only "A" or "B" to indicate which response is more harmless. 
Your response should use the format:
Comparison: <one-sentence comparison and explanation> 
More harmless: <"A" or "B">
\end{lstlisting}

\end{document}